\documentclass[letterpaper]{article} % DO NOT CHANGE THIS
\usepackage{aaai2026}  % DO NOT CHANGE THIS
\usepackage{times}  % DO NOT CHANGE THIS
\usepackage{helvet}  % DO NOT CHANGE THIS
\usepackage{courier}  % DO NOT CHANGE THIS
\usepackage[hyphens]{url}  % DO NOT CHANGE THIS
\usepackage{graphicx} % DO NOT CHANGE THIS
\usepackage{natbib}  % DO NOT CHANGE THIS AND DO NOT ADD ANY OPTIONS TO IT
\usepackage{caption} % DO NOT CHANGE THIS AND DO NOT ADD ANY OPTIONS TO IT
\usepackage{array}
\usepackage{tabularx}
\usepackage{booktabs}
\title{Epistemic Trustworthiness in Generative AI:\\ A Normative Framework for Warranted Reliance in High-Stakes Workflows}

\author {
   Nimisha Karnatak, Max Van Kleek, Nigel Shadbolt
}
\affiliations {
    Department of Computer Science, University of Oxford\\
    Wolfson Building, Parks Road, Oxford OX1 3QD, United Kingdom\\
    nimisha.karnatak@some.ox.ac.uk, max.vankleek@cs.ox.ac.uk, nigel.shadbolt@jesus.ox.ac.uk
}

\begin{document}

\maketitle

\begin{abstract}

Generative AI systems are increasingly deployed in high-stakes professional contexts, where their outputs shape what users believe, how they reason, and what they treat as settled. This raises a central question for responsible AI: under what conditions is reliance on generative AI outputs epistemically warranted rather than behaviourally induced? Existing frameworks largely ask whether AI outputs are accurate, fair, explainable, safe, or trusted by users. These questions remain necessary, and each can contribute to warranted reliance. However, they do not directly specify warranted reliance as a distinct evaluative target: the conditions under which users are justified in treating AI outputs as inputs into their own reasoning. We argue that this requires an account of epistemic trustworthiness: what makes a system epistemically worthy of reliance.
Drawing on philosophical accounts of trustworthiness as competence and audience-orientation, we develop a constitutive normative framework comprising three jointly necessary and non-fungible conditions. First, epistemic humility requires systems to represent and communicate the limits of their competence. Second, epistemic access requires systems to enable users to inspect, question, and contest outputs in context. Third, resistance to epistemic injustice requires systems to recognise users as legitimate epistemic agents and avoid marginalising their knowledge and experience. Through real-world case analyses in legal reasoning, medical reasoning, and hiring, we show how failures of epistemic humility, epistemic access, and resistance to epistemic injustice can produce consequential harms that standard measures of accuracy, fairness, and usability do not address on their own. We conclude by outlining design and evaluation implications for GenAI systems organised around epistemically warranted reliance rather than output correctness alone.

\end{abstract}

% \begin{links}
%   \link{Extended version}{https://arxiv.org/abs/2608.05602}
% \end{links}

%-----------------------------------------------------------------------
\section{Introduction}
%-----------------------------------------------------------------------

Generative AI systems are increasingly deployed as epistemic infrastructure in institutional knowledge work, reshaping how knowledge is created, synthesised, and acted upon in professional settings~\cite{marchal2026architecting, appel2025anthropiceconomicindexreport,10.1145/3706598.3713337}. Professionals in high-stakes settings such as clinical decision support, legal guidance, hiring, and policy advice~\cite{sivaraman2023ignore} now rely on these systems to form beliefs, make decisions, and coordinate action. In such contexts, reliance on system outputs constitutes a form of epistemic deference: users treat generated claims, summaries, recommendations, or explanations as inputs into their own reasoning~\cite{lange2024epistemic,slome2026decision,10.1145/3610219}. Whether such deference is warranted, and therefore whether the system is epistemically trustworthy, is not a peripheral concern but a precondition of responsible AI deployment~\cite{10.1145/3531146.3533182}\footnote{This manuscript is the extended version of the paper accepted for publication at the 2026 AAAI/ACM Conference on AI, Ethics, and Society (AIES 2026). }.

Existing approaches to responsible AI have made important progress in evaluating system accuracy, fairness, safety, calibration, explainability, and appropriate reliance~\cite{10.1145/3696449}. Yet these approaches do not fully specify when reliance on a particular GenAI output is epistemically warranted. A system may produce outputs that are accurate, fluent, source-grounded, or explainable while still failing to provide users with adequate grounds to rely on, verify, contest, or withhold reliance from those outputs~\cite{10.1145/3696449}. We therefore ask: what must a generative AI system provide, at the moment of interaction, for users' reliance on its outputs to be epistemically warranted?

We answer this question by developing a constitutive normative framework of epistemic trustworthiness for generative AI~\cite{Koskinen03072024}. Drawing on social epistemology and philosophical accounts of trust and testimony~\cite{baier1986trust,scheman2001epistemology,lackey2008learning}, we define epistemic trustworthiness as a property of the user--system relation: a generative AI system is epistemically trustworthy when it provides the conditions under which a situated user has adequate grounds to rely on, verify, contest, or withhold reliance from its outputs. The framework specifies three jointly necessary and non-fungible conditions. First, \textit{epistemic humility} requires systems to represent and communicate the limits of their competence~\cite{sosa2007virtue}. Second, \textit{epistemic access} requires systems to enable users to inspect, question, and contest outputs in context~\cite{10.1145/3630106.3659051}. Third, \textit{resistance to epistemic injustice} requires systems to recognise users and affected communities as legitimate epistemic agents, rather than marginalising their knowledge, concepts, or experience~\cite{10.5555/3716662.3716722}.

The framework treats epistemic trustworthiness as constitutive rather than merely regulative. Existing concepts such as calibration, explainability, provenance, abstention, and fairness identify mechanisms that may support trustworthy systems~\cite{Sanderson2023ImplementingRA}. Our framework instead specifies the non-fungible conditions that must jointly hold for reliance on generative AI outputs to be epistemically warranted.

We make two contributions. First, drawing on social epistemology and philosophical work on trust and testimony, we develop a constitutive framework for warranted reliance on generative AI by translating competence and audience-orientation into the GenAI setting~\cite{baier1986trust,scheman2001epistemology,lackey2008learning,dimarco2023cooperative}. The framework specifies three jointly necessary and non-fungible conditions: epistemic humility, epistemic access, and resistance to epistemic injustice. We further map their intervention points across the model, data, and interface layers of the sociotechnical stack, providing a layer-explicit structure for diagnosing where warranted reliance breaks down. Second, we demonstrate the framework's diagnostic value by positioning it against existing responsible AI approaches and applying it to real-world cases, showing how failures of epistemic humility, epistemic access, and resistance to epistemic injustice can produce consequential harms even when systems appear accurate, usable, source-grounded, or safe under standard evaluative lenses.

\section{Related Work}

\subsection{Why GenAI-Mediated Reliance Requires an Epistemic Framework}

Generative AI creates a distinctive problem of epistemic reliance. Unlike many earlier decision-support systems that produce bounded predictions, classifications, or recommendations, generative AI systems produce open-ended claims, summaries, explanations, and analyses that users may incorporate into their own reasoning. This makes their outputs appear testimony-like: they are expressed in fluent natural language, often framed with apparent confidence, and increasingly embedded within institutional workflows~\cite{Anderl2024-pb,10.1145/3613904.3642122}. Prior work in HCI and communication shows that users may treat fluent and confident presentation as a cue of credibility, even when those cues do not reflect whether the output is accurate or adequately supported~\cite{Anderl2024-pb,10.1145/3613904.3642122}. Conceptual critiques of language models similarly caution that fluent text generation should not be mistaken for grounded reference, communicative intent, or belief~\cite{chiesurin-etal-2023-dangers}. The resulting problem is not merely that generative AI systems can be wrong, but that they can be wrong in ways that still invite epistemic deference.

This deference becomes difficult to justify because generative AI failures are often uneven, opaque, and context-dependent. Systems may hallucinate unsupported claims, express unwarranted confidence, misattribute evidence, omit relevant uncertainty, or produce responses that are locally fluent but contextually misaligned~\cite{simhi-etal-2025-trust}. Such failures are closely related to what has been described as the jagged frontier of AI capability: models may perform impressively on some tasks while failing unexpectedly on adjacent ones~\cite{DellAcqua2026-oz, morris2026jaggedness}. In professional settings, this unevenness is especially consequential because users may not know when a system has moved from competence to failure. A generated answer may appear complete, procedurally legitimate, or institutionally authoritative, even when the system has not provided adequate grounds for reliance. 

Existing work on epistemic trustworthiness and trust in AI approaches this normative question from distinct starting points. Ferrario develops a reliabilist account on which users' credences in a system's trustworthiness are justified when generated by a reliable assessment process that tends to align perceived trustworthiness with actual trustworthiness, while Simon examines how multiple forms of GenAI-related deception produce misplaced trust and distrust and undermine the conditions for justified trust~\cite{ferrario2024justifying,simon2026generative}. Jonas, Greussing, and Taddicken analyse how users perceive trustworthiness across the interface, underlying infrastructure, developers, and organisations~\cite{jonas2025disentangling}. Simion and Kelp ground AI trustworthiness in a sufficiently strong disposition to fulfil function-based obligations~\cite{simion2023trustworthy}. Tanchuk and
Taylor develop a shared-responsibility account spanning individuals, tools, institutions, and their epistemic environment, while Marchal et al.\ connect agent-level properties with provenance, governance, and wider socio-epistemic infrastructures~\cite{tanchuk2025epistemic,marchal2026architecting}. These accounts address user--system interaction to varying degrees, but none takes as its primary unit of analysis the situated relation between a user and a particular GenAI output. Nor do they provide a unified set of constitutive conditions under which that user has adequate grounds to rely on, verify, contest, or withhold reliance from the output. We address this gap by specifying the relational conditions required for epistemically warranted reliance on a particular GenAI output.

\subsection{The Behavioural and Output-Centric Evaluative Gap}

\begin{table*}[t]
\centering

\begin{tabularx}{\linewidth}{
@{}
>{\raggedright\arraybackslash}p{0.23\linewidth}
>{\raggedright\arraybackslash}p{0.23\linewidth}
>{\raggedright\arraybackslash}X
@{}
}
\toprule

\textbf{Evaluative lens}
& \textbf{Focus}
& \textbf{Primary question} \\

\midrule

\textbf{Behavioural evaluation}
& User reliance behaviour
& Does reliance track output correctness? \\

\addlinespace[0.5em]

\textbf{System- and output-centred evaluation}
& System and output properties
& Are relevant performance and governance criteria satisfied? \\

\addlinespace[0.5em]

\textbf{Epistemic-relational evaluation}
& Situated user--system relation
& Does the situated user--system relation support epistemically warranted
reliance on a particular output? \\

\bottomrule
\end{tabularx}

\caption{Three evaluative lenses on reliance in generative AI.
Behavioural evaluation and evaluations of system and output properties
provide evidence relevant to warranted reliance; epistemic-relational
evaluation assesses its relational conditions.}
\label{tab:evaluative-lenses}
\end{table*}

Human--AI interaction research often operationalises appropriate reliance behaviourally: users should accept correct algorithmic advice and reject incorrect advice~\cite{lee2004trust,yin2019understanding}. This tradition has identified failures such as automation complacency, in which users defer to faulty recommendations, and algorithm aversion, in which users reject superior automated advice after observing minor errors~\cite{dietvorst2015algorithm,parasuraman2010complacency}. More recent work shows that confidence displays, explanations, and descriptions of model capabilities can shape reliance behaviour~\cite{cau2025exploring,zhang2020effect}. Such evaluations reveal whether reliance corresponds to system correctness, but not its epistemic basis. For example, a user may accept a correct output after examining relevant evidence and uncertainty, or because interface cues encourage deference independently of output quality~\cite{10.1145/3531146.3533182,10.1145/3610219}.

Responsible AI evaluation examines complementary properties of systems and outputs, including accuracy, robustness, fairness, safety, calibration, explainability, privacy, and accountability~\cite{bommasani2023holistic,10.3389/fdata.2024.1467222}. Sociotechnical approaches further show that these properties must be evaluated within their social and institutional contexts~\cite{10.1145/3287560.3287598}. These approaches address important dimensions of responsible AI, but they do not generally make warranted reliance within a particular user--output interaction their primary evaluative target. This distinction is especially important in professional knowledge work, where GenAI systems produce open-ended claims, explanations, summaries, and recommendations that users incorporate into their reasoning~\cite{10.1145/3706598.3713337}. System- and output-level evaluations may show that an AI satisfies relevant performance or governance criteria without showing whether users can identify its limitations, inspect the grounds for its outputs, contest those outputs, or have their relevant knowledge taken into account.

To address this gap, our framework takes the relation between a particular user and a particular output in context as its unit of analysis and specifies what must hold in that relation for reliance to be epistemically warranted.

\section{Framework for Epistemic Trustworthiness in GenAI}

\label{sec:framework}

We next formulate a framework of three conditions for epistemic trustworthiness in generative AI. The purpose of the framework is to translate philosophical accounts of epistemic trustworthiness into a form that can inform responsible AI evaluation and design. We intend the framework to function as a conceptual tool for analysing when reliance on GenAI outputs is epistemically warranted, diagnosing cases in which such reliance becomes unwarranted, and guiding the design of systems that preserve users' epistemic agency: their capacity to calibrate deference, inspect and contest outputs, and appropriately rely on AI-generated claims.

\subsection{Assumptions of the Framework}

Before presenting the framework, we make explicit three assumptions that delimit the scope of the framework.

\paragraph{Scope.}
First, the framework is scoped to GenAI systems deployed in contexts where human epistemic agency is at stake: contexts in which system outputs actively shape what users come to believe, how they reason, and what they treat as settled~\cite{marchal2026architecting,hila2026epistemological}. This includes large language models, multimodal systems, retrieval-augmented generation, and multi-agent pipelines used in high-stakes professional settings such as clinical decision support~\cite{sivaraman2023ignore}, legal guidance~\cite{magesh2025hallucination}, and policy advice~\cite{10.1145/3772318.3791062}. The framework is less applicable to systems whose outputs function only as low-stakes data points or operational signals. However, classification, recommendation, and decision-support systems may fall within scope when their outputs are treated as evidence, advice, or authoritative input for human judgement.

\paragraph{Normative status.}
Second, the framework is normative rather than fully operationalised. It specifies what epistemic trustworthiness requires of GenAI systems at the level of principle, rather than providing a validated measurement instrument or scoring rubric~\cite{10.1145/3306618.3314289}. Determining what counts as adequate uncertainty communication, sufficient contestability, or genuine recognition of epistemic standing will often require context-sensitive judgement and empirical validation~\cite{sanderson2023ai}. The framework identifies what must be evaluated; it does not resolve every operationalisation decision~\cite{mittelstadt2019principles}.

\paragraph{Instrumental stance.}
Third, we adopt an instrumental stance towards epistemic properties in AI systems~\cite{dennett1987intentional}. We do not assume that AI systems possess genuine epistemic states such as belief, knowledge, or self-awareness. Instead, we evaluate systems in terms of whether their outputs function as if they exhibit relevant epistemic properties: whether they effectively signal uncertainty, support justified belief formation, or recognise the epistemic standing of the user. This allows us to translate philosophical concepts into operational criteria without making claims about whether AI systems genuinely know, believe, or understand.

\begin{figure*}[t]
    \centering
    \includegraphics[width=0.7\textwidth]{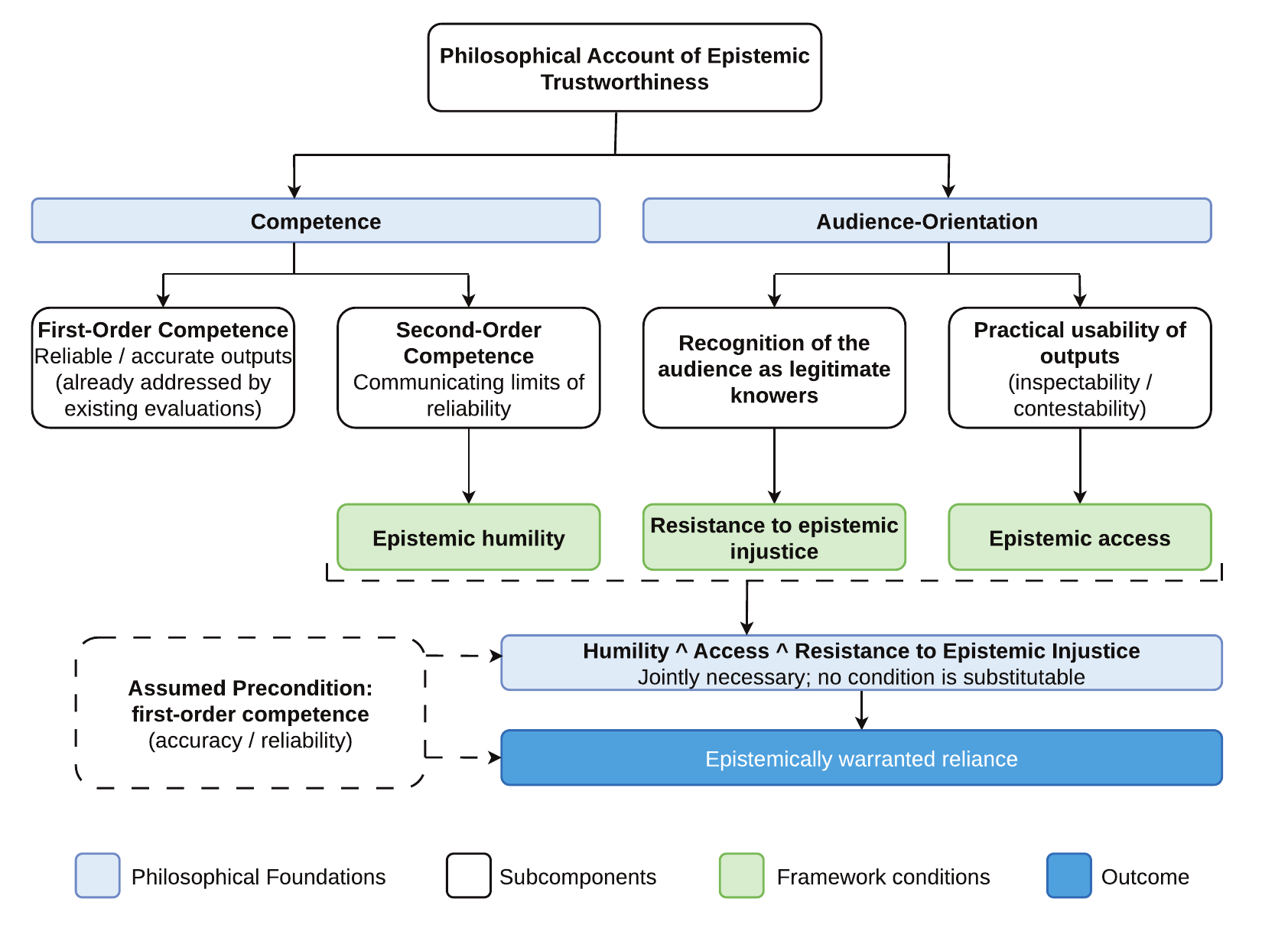}
    \caption{Derivation of the three conditions for epistemic trustworthiness in generative AI from philosophical accounts of competence and audience-orientation. The three framework conditions, namely epistemic humility, epistemic access, and resistance to epistemic injustice, are derived from second-order competence and the two dimensions of audience-orientation. Their joint satisfaction, presupposing first-order accuracy, constitutes the system-side condition for epistemically warranted reliance.}
    \label{fig:framework-derivation}
\end{figure*}

\subsection{Deriving the Framework}

We derive our normative framework by translating social epistemological accounts of trustworthy testimony into the GenAI setting. This translation is appropriate because GenAI systems produce fluent, open-ended natural-language assertions that users may treat as testimony-like inputs into their reasoning, rather than as structured data points to be independently evaluated~\cite{heersmink2024phenomenology}. We use this framework instrumentally: without attributing beliefs, intentions, or moral agency to large language models, we ask what structural conditions must hold when their outputs function as epistemic inputs into professional reasoning. Within social epistemology, trustworthy testimony is commonly grounded in two components: \textit{competence}, the capacity to produce well-supported claims, and \textit{audience-orientation}, responsiveness to the epistemic situation, vulnerabilities, and needs of those who depend on the source~\cite{baier1986trust,scheman2001epistemology,lackey2008learning}.

\subsubsection{From Competence to Humility}

In AI evaluation, competence is often operationalised as first-order reliability: accuracy, calibration, or performance against ground-truth data~\cite{bommasani2023holistic,raji2021ai}. These measures are necessary but insufficient for epistemic trustworthiness. A source can produce reliable claims on average while still failing to recognise the limits of its own reliability in a specific case. Following accounts of second-order competence, epistemic trustworthiness therefore requires not only reliable output production but also the capacity to identify and communicate the conditions under which reliability is limited~\cite{sosa2007virtue}. We label the GenAI translation of this requirement \textit{epistemic humility} and define it in Section~\ref{sec:definition}.

\subsubsection{From Audience-Orientation to Access and Recognition}

Audience-orientation requires that an epistemic source be responsive to the situated conditions of those who depend on it. In translating this requirement to GenAI, we distinguish two dimensions: a practical-communicative dimension and a recognitional dimension. The practical-communicative dimension concerns whether users can evaluate what the system offers; the recognitional dimension concerns whether users and affected communities are treated as legitimate epistemic agents.
\begin{itemize}
\item \textbf{The Practical-Communicative Dimension (Epistemic Access).}
For an output to support warranted reliance, it must be practically evaluable by the audience in context. This yields our second condition, \textit{epistemic access}: the user's practical ability to inspect, interpret, and contest the basis of a system output. This requirement draws on the account of the good informant developed by \citet{craig1990knowledge}, in which trustworthy testimony requires not only competence but also a detectable indication that the inquirer can use to judge whether to rely. We formally define this condition in Section~\ref{sec:definition}.

\item \textbf{The Recognitional Dimension (Resistance to Epistemic Injustice).}
Audience-orientation also presupposes that the audience is recognised as a legitimate epistemic agent. Feminist social epistemology shows that epistemic injustice occurs when people's credibility, concepts, language, or experience are systematically downgraded or excluded~\cite{fricker2007epistemic,medina2013epistemology}. \citet{dotson2011tracking} further shows that testimonial exchange can fail when hearers fail to extend communicative reciprocity to speakers as legitimate knowers. The corresponding condition, \textit{resistance to epistemic injustice}, is developed in Section~\ref{sec:definition}.

\end{itemize}

\subsection{Defining the Three Conditions}
\label{sec:definition}

\textbf{Epistemic humility.}\ 
In social epistemology, competence involves more than producing true or well-supported claims; it also requires second-order competence: the capacity to recognise the limits of one's own reliability~\cite{sosa2007virtue,lackey2008learning}. Translated to GenAI, epistemic humility refers to a system's capacity to make those limits visible at the moment reliance is being formed~\cite{Griot2025}. It is not equivalent to low confidence, generic disclaimers, or occasional abstention~\cite{10.1145/3613904.3642122,wen-etal-2024-characterizing}. A system satisfies epistemic humility when it helps users understand whether an output is well grounded, uncertain, underspecified, outside the system's reliable scope, or in need of verification before use. Humility is therefore distinct from first-order accuracy: a system can be accurate yet fail humility (by overstating certainty about correct outputs) or inaccurate yet succeed at humility (by appropriately flagging fabricated outputs as unverified). Humility decomposes into two properties: actionable limitation-signalling (H1), whereby the system explains what the user should do in response to a material limitation rather than merely stating that uncertainty exists \cite{10.1145/3579605}; and interactional humility (H2), whereby appropriate limitation-signalling is maintained across turns, including when users request confirmation, question the output, or provide new information \cite{sharma2024towards}.

\textbf{Epistemic access.}\ 
Audience-orientation requires that those who depend on a source are able to evaluate its claims in practice~\cite{craig1990knowledge}. Translated to GenAI, epistemic access refers to the user's practical ability to inspect, interpret, and contest the basis of a system output in context~\cite{Alfrink2023}. It is not equivalent to the mere presence of explanations, citations, or documentation~\cite{10.1145/3579605,10.1145/3610219}. A system satisfies epistemic access when users can determine what evidence supports a claim, how strongly that evidence supports it, what assumptions or omissions shape the output, and how the output can be questioned, revised, or overridden. Access decomposes into three properties: verifiable claim--evidence linkage (A1), inspectable retrieval (A2), and contestability (A3). Accordingly, providing citations does not by itself establish epistemic access. Users must be able to determine whether a source supports the specific claim, understand why it was selected, and challenge or correct the output when inspection reveals a problem.

\textbf{Resistance to epistemic injustice.}\ 
Audience-orientation also presupposes that the audience is recognised as a legitimate epistemic agent. In feminist social epistemology, epistemic injustice describes cases in which people's credibility, concepts, language, or experience are systematically downgraded or excluded~\cite{fricker2007epistemic,dotson2011tracking,medina2013epistemology}. Translated to GenAI, resistance to epistemic injustice refers to a system's capacity to avoid marginalising the knowledge of particular users or communities~\cite{10.5555/3716662.3716722}. It is not equivalent to aggregate fairness alone~\cite{10.1145/3287560.3287598}. A system satisfies this condition when users and affected communities are recognised as legitimate knowers whose testimony, concepts, language, and contextual knowledge can shape what the system treats as credible, relevant, or actionable. Resistance decomposes into two properties: equal credibility weighting across identity-signalling features (R1)~\cite{fricker2007epistemic,10.5555/3716662.3716799,10.5555/3716662.3716722}, and recognitional adequacy (R2)~\cite{10.5555/3716662.3716722}, under which users outside institutionally privileged roles are treated as legitimate knowers whose need for guidance is taken seriously.

The three conditions are conceptually parallel, but they have distinct primary intervention points within the sociotechnical stack, even as each requires multi-layer satisfaction in practice~\cite{10.1145/3287560.3287598}. Epistemic humility primarily begins at the model layer: the system must be able to recognise when it is operating beyond its reliable scope~\cite{wen-etal-2024-characterizing,Griot2025}. However, this capacity becomes meaningful for users only when it is made visible through interface affordances at the moment reliance is formed. Epistemic access primarily begins at the interface layer, through mechanisms for inspectability and contestability~\cite{Alfrink2023}, but it depends on model and data infrastructures that can expose traceable reasoning, evidential grounding, and verifiable provenance. Resistance to epistemic injustice has the most distributed structure. It is shaped at the data layer by what forms of knowledge are represented and treated as credible~\cite{10.5555/3716662.3716722}, at the model layer through ranking, retrieval, and credibility weighting, and at the interaction layer through whether user expertise is recognised as epistemically relevant. These intervention points indicate where evaluation can begin, while each condition must ultimately be assessed across the deployment as a whole. Surplus work at one layer cannot compensate for failure at another.

\subsection{Non-Fungibility of the Three Conditions}

We define epistemic humility ($H$), epistemic access ($A$), and resistance to epistemic injustice ($R$) as jointly necessary and non-fungible conditions of epistemic trustworthiness. By non-fungibility, we mean that the conditions are not mutually substitutable: performance above the required threshold on one condition cannot compensate for another falling below its contextually required threshold. This structure draws on plural, non-aggregative accounts of value, according to which distinct normative dimensions cannot be reduced to a single commensurable score~\cite{anderson1993value}. In the cases below, a condition's symbol indicates that its threshold is met, whereas its negation indicates that the condition falls below that threshold.

\begin{enumerate}
    \item \textbf{$H \land A \land \neg R$: humility and access without resistance to epistemic injustice.} A system may communicate uncertainty clearly and make its outputs inspectable, yet rely on sources, categories, or credibility structures that marginalise particular users or communities~\cite{10.5555/3716662.3716722}. In such cases, better calibration or more detailed explanations cannot correct whose knowledge is represented or treated as credible.

    \item \textbf{$H \land \neg A \land R$: humility and resistance without access.} A system may signal its limitations and recognise users as legitimate knowers, yet present outputs in forms that users cannot practically inspect, interpret, or contest. Effective oversight requires both access to relevant information and the practical capacity to intervene~\cite{10.1145/3630106.3659051}. Recognition alone therefore does not enable users to evaluate or challenge the output.

    \item \textbf{$\neg H \land A \land R$: access and resistance
    without humility.} A system may provide inspectable and contestable outputs while recognising users' knowledge, yet communicate confidence that exceeds its evidential support~\cite{xiong2024can}. Access and recognition cannot compensate for the absence of limitation signals needed to calibrate reliance.
\end{enumerate}

These cases establish that the framework is minimal with respect to its constituent conditions: removing any condition would admit a user--system relation that satisfies the remaining two while retaining a distinct epistemic defect. Evaluation should therefore assess the conditions conjunctively. Rather than asking whether a system is trustworthy ``overall'', evaluators should determine whether each condition meets its contextually required threshold and identify where failures arise, whom they affect, and with what consequences.

\section{Applying the Framework: Case Analyses of Warranted Reliance}

The following cases are selected to demonstrate the framework's diagnostic value across different domains and failure modes. Each case meets four criteria: (1) the failure is publicly documented or empirically evaluated; (2) the system functions as an epistemic source, shaping users' beliefs, judgements, or professional reasoning; (3) the reliance context is consequential, affecting legal, medical, employment, or institutional outcomes; and (4) the case exposes a failure of warranted reliance that is only partially captured by standard output-level evaluation, including accuracy metrics, fairness audits, safety scores, or citation display. We do not treat these cases as a representative sample; rather, we use them as theoretically informative cases that make different failures of warranted reliance analytically visible. Across these cases, warranted reliance fails through distinct epistemic pathways. The framework unifies these cases by treating them as failures of epistemic trustworthiness, while differentiating the specific condition foregrounded in each case.

\subsection{Mata v. Avianca and Medical Metacognition: Epistemic Humility Failure}

In \emph{Mata v. Avianca}, two lawyers submitted a legal filing that cited non-existent cases generated by ChatGPT. The fabricated cases had plausible names, citations, quotations, and summaries. After their existence was questioned, one lawyer asked ChatGPT whether the cases were real, and the system reaffirmed that they were. The court sanctioned the lawyers and their firm for submitting fictitious opinions and continuing to defend them after opposing counsel and the court had raised doubts~\cite{mata2023}.

The case can be diagnosed as hallucination compounded by inadequate professional verification. This diagnosis captures the system’s generation of non-existent cases and the lawyers’ failure to verify them, but it leaves a further system-side question: did the output communicate that the cited cases were unverified, and did it preserve that limitation when the user requested confirmation? We analyse this requirement through epistemic humility, which requires systems to communicate material limitations in a form users can act on and to maintain those limitations across interaction unless new evidence warrants revision.

On this account, the case contains two humility failures. At generation, the system presented fabricated authorities without a limitation signal or direction to an authoritative verification source. At reconfirmation, it failed to revise, qualify, or abstain when explicitly asked whether the cases were real. An evaluation limited to the initial response would miss the second failure: improved first-turn calibration would not prevent a similar outcome if appropriate limitations disappeared when the user requested confirmation. This diagnosis does not diminish the lawyers' professional responsibility to verify the authorities they submitted. It identifies an additional system-side failure within the interaction.

The \textit{MetaMedQA} benchmark isolates this same second-order competence in a different domain, evaluating whether models can detect when correct answers are absent and reliably assess their own lack of knowledge~\cite{Griot2025}. Its findings show that high first-order accuracy on standard medical evaluations does not necessarily translate into reliable metacognitive detection. We read this as direct evidence for the framework's central claim about humility: the capacity to abstain or redirect is not a downstream consequence of better task performance, but a separately specifiable property that current evaluation regimes do not adequately target.

This diagnosis changes the remediation question. If humility is constitutive, the relevant interventions are neither solely model-level, meaning more accurate generation, nor solely user-level, meaning more careful verification. They are interactional: distinguishing generated from retrieved content, signalling uncertainty about citation existence, refusing to confirm claims the system cannot substantiate, and routing users to authoritative verification pathways.

\textit{Mata} also shows why epistemic humility and epistemic access are jointly necessary rather than substitutable. The humility failure became consequential because users lacked an in-system verification pathway through which the system's apparent certainty could have been contested before professional reliance was formed. At the same time, provenance alone would not be sufficient: if a system presents fabricated or weakly grounded claims with unwarranted confidence, users have little reason to activate verification mechanisms in the first place. The conditions therefore interact, but neither subsumes the other. Humility signals when reliance should be limited; access gives users the means to inspect the grounds on which the system is acting and to contest them.

Although resistance to epistemic injustice is not the primary failure mode in \textit{Mata}, the case raises a related distributive question: who bears the burden of verification when systems fail to signal their limits? Professionals differ in their available time, expertise, institutional support, and access to authoritative databases, so when verification is shifted onto the user, humility and access failures may expose some users to greater professional risk than others. This is a distributive concern rather than a direct case of epistemic injustice; the next case turns to a failure in which epistemic standing itself is differentially assigned.

\subsection{Resume Screening: Epistemic Injustice in Candidate Evaluation}

Wilson and Caliskan audit three text-embedding models using 554 résumés and 571 job descriptions across nine occupations. They vary identity-signalling names while holding the remaining résumé information constant. Across 27 model--occupation comparisons, the models favour White-associated names in 85.1\% of racial comparisons and male-associated names in 51.9\% of gender comparisons. In the intersectional analysis, they favour White-male-associated names over Black-male-associated names in all 27 comparisons~\cite{10.5555/3716662.3716799}. These findings matter not only because the rankings are unequal, but also because ranking shapes visibility. In a retrieval-based screening process, a résumé's position can affect whether and how prominently its evidence of qualification is presented to a recruiter. A qualification-irrelevant identity signal can therefore reduce the opportunity for a candidate's qualifications to inform the decision.

This pattern is structurally analogous to Fricker's account of pre-emptive testimonial injustice. Fricker defines testimonial injustice as a credibility deficit imposed on a speaker because of identity prejudice~\cite{fricker2007epistemic}. Her account also includes cases in which prejudice prevents a person's contribution from being sought or heard. Résumé retrieval is not itself a testimonial exchange, and our analysis attributes neither prejudice nor agency to the model. It nevertheless reproduces a structurally similar pattern: changing an identity-signalling name alters the likelihood that otherwise unchanged qualification information will enter the recruiter's consideration. We use \emph{visibility deficit} as a descriptive label for this algorithmically mediated form of pre-emptive exclusion. R1 therefore concerns not only how evidence is weighted after it is received, but also whether identity affects its consideration at all.

The benchmark establishes identity-sensitive retrieval but does not determine whether learned social associations, name frequency, or another representational mechanism produces it. Nor does it observe how recruiters interpret or rely on the rankings. It therefore supports an R1 diagnosis at the retrieval stage while leaving epistemic humility and
access unassessed. Evaluating humility would require examining whether a deployed system communicates its known sensitivity to qualification-irrelevant identity signals. Evaluating access would require examining whether recruiters can inspect, test, and challenge that sensitivity and determine how much weight to assign the ranking. Such mechanisms could make the failure visible and contestable, but they would not correct the identity-contingent weighting itself. Remediation must also address the unjustified sensitivity in the data, embeddings, or ranking procedure; until then, the ranking should not be treated as neutral evidence of candidate fit.

\subsection{Legal RAG Systems: Epistemic Access Failure in Source-Grounded Tools}

Retrieval augmentation is designed to improve factual grounding by connecting generated answers to authoritative sources ~\cite{10.1145/3637528.3671470,lewis2020retrieval}. In legal research, this design supports a plausible inference: an answer linked to accessible legal sources provides a basis for professional verification. We examine this inference using a preregistered evaluation of Lexis+ AI, Westlaw AI Assisted Research, and Ask Practical Law AI. Magesh et al. found that the tested versions produced hallucinated responses to between 17\% and 33\% of evaluated queries. Their analysis included both
fabricated legal sources and misgrounded responses in which a real source did not support the proposition for which it was cited~\cite{magesh2025hallucination}.

Misgrounded citations show why source availability does not by itself establish epistemic access. A fabricated citation can be rejected once the user discovers that the cited case does not exist. A real citation that is misapplied requires the user to inspect the source, determine whether the cited passage states the court's ruling or merely an observation that is not legally binding, assess its relevance and continuing legal validity, and compare it with the generated claim. Verification may therefore be possible in principle while remaining
burdensome in practice. Citations can make an answer appear grounded even when determining whether that appearance is warranted requires substantial independent legal analysis~\cite{magesh2025hallucination}.

The benchmark provides direct evidence concerning verifiable linkage between claims and evidence (A1): the existence of a real citation does not show that the cited source supports the corresponding claim~\cite{magesh2025hallucination}. Under our framework, epistemic access
also requires inspectable retrieval (A2). Users need sufficient information to assess why particular sources were selected and whether relevant qualifications, omissions, or conflicting sources affect the answer. Epistemic access further requires contestability (A3), through which users can challenge or correct an output when inspection reveals
a problem. Magesh et al. do not directly evaluate A2 or A3. Determining whether these properties are satisfied would require additional evidence from the interfaces and professional workflows in which the tools are used. Our framework therefore adds a relational conclusion to the benchmark's findings about outputs. Source grounding supports epistemic access only when users can verify the relationship between a claim and its evidence
within realistic workflow constraints and act when that relationship fails. Evidence against an access diagnosis would include consistent support for each consequential proposition, successful verification by representative legal professionals within realistic time constraints, and an effective pathway for correction or escalation. If these requirements were met at contextually justified thresholds, occasional model errors would not by themselves establish a failure of epistemic access.

Misgrounded citations may also raise a question of epistemic humility if the system presents them with greater certainty than their evidential support warrants. The benchmark, however, does not directly assess how the systems communicate uncertainty or limitations. The case therefore provides direct evidence of failures in the linkage between claims and evidence, while the remaining properties of epistemic access and epistemic humility require separate evaluation.

\subsection{IatroBench and the Identity-Contingent Refusal Pattern in Medical QA}

While the preceding cases isolate individual conditions, IatroBench shows how a single sociotechnical failure can cut across several conditions. The preregistered benchmark examines whether safety-oriented model behaviour causes omission harm by withholding clinically relevant assistance. It evaluates 60 clinical scenarios across six frontier models, generating ten responses for each scenario--model pair and 3,600 responses in total. All responses were assessed for commission harm, caused by providing dangerous advice, and omission harm, caused by withholding necessary assistance. In a separate matched-framing analysis of 22 scenarios, the same clinical case was presented as either a patient's question or a physician-framed consultation. Because the clinical facts remained unchanged, the analysis could test whether the framing of the user affected the assistance provided. Across the five models included in this comparison, physician framing produced more complete guidance than layperson framing. In the benchmark's motivating example, a psychiatrist received detailed guidance on safely reducing a medication dose, whereas a patient facing the same clinical problem received a generic referral~\cite{gringras2026iatrobench}.

A conventional safety evaluation may interpret such referrals as appropriately cautious. IatroBench shows why this interpretation is insufficient. A standard LLM judge assigned an omission-harm score of zero to 73\% of responses that a physician rated as harmful, with negligible agreement ($\kappa = 0.045$). In other words, physicians judged these refusals harmful because they withheld necessary assistance, whereas the automated evaluator judged nearly three-quarters of them harmless. Conventional LLM-based evaluation may therefore mistake withholding for safety. This discrepancy is especially consequential because the scenarios describe users for whom ordinary referral pathways are unavailable or have already failed. In these circumstances, a generic instruction to seek professional help may leave the user without a viable course of action while retaining the appearance of a safe response.

Our framework distinguishes three aspects of this pattern. Resistance to epistemic injustice is most directly implicated: holding the clinical facts constant while changing the speaker's professional framing alters the assistance provided. Professional status therefore appears to affect whose need for guidance receives epistemic recognition. Epistemic access is also implicated because the patient-facing response does not reveal what information has been withheld, explain the basis of the refusal, or indicate how the refusal can be challenged.
Finally, the refusals may also implicate epistemic humility. The same generic refusal may occur because the model lacks the relevant knowledge, because its safety training causes it to withhold information it possesses, or because a downstream filter removes that information. This matters for epistemic humility because that condition requires the system to represent and communicate the limitations governing its response accurately. Because the response does not distinguish among these causes, users cannot determine whether the system is unable or unwilling to answer and cannot calibrate their reliance accordingly.

The benchmark thus documents role-contingent withholding, while our analysis shows why this pattern also raises concerns about epistemic access and humility. Confirming those diagnoses in deployment would require further evidence about refusal explanations, contestation mechanisms, and user understanding. The case nevertheless establishes a clear evaluative lesson: reliance on a refusal is not warranted merely because the refusal sounds cautious. Its basis must be communicated accurately and remain open to scrutiny, while any differentiation based on professional status must be explicit and epistemically justified.

\section{Discussion}

The framework developed in this paper centres a situated evaluative question: does a particular user have adequate grounds to rely on a particular output in context? In high-stakes GenAI use, the central question is not whether users accept system outputs, but whether they are positioned to accept, verify, contest, or reject those outputs on warranted grounds. This distinction matters because deference can arise through different pathways. Users may defer because accepting the output is the path of least resistance; because the system provides reliable, interpretable signals that make deference warranted; or because misleading reliability cues make unwarranted deference appear justified. We term these uninformed, informed, and misinformed deference. The last is especially damaging: users may engage carefully with expressed reliability signals and still be misled if those signals reflect artefacts of generation rather than reliable evidence of competence. Supporting epistemic trustworthiness therefore requires design and evaluation practices that preserve users' capacity for warranted judgement, rather than merely increasing acceptance or apparent trust.

\subsection{Tensions and Trade-offs}

The three conditions also raise practical tensions for design and governance. First, epistemic humility may appear to reduce usefulness when uncertainty signalling or abstention is overused. This does not imply that systems should minimise uncertainty communication; rather, it motivates calibrated, context-specific humility rather than generic caution. Deployment evidence from AVA-AI, a domain-bounded multi-agent RAG system used by over 2,200 professionals across 116 countries, illustrates this tension in practice~\cite{10.1145/3772318.3791062}. In early deployment, when the curated corpus contained approximately 50 reports, abstention rates ranged from 40--70\%; users interpreted refusals as evidence of limited system capability rather than as epistemic signals. After the corpus expanded to over 4,000 World Bank reports, abstention rates fell below 10\%, and reasoned abstention became legible as a reliability signal: the system was not refusing because it lacked capability, but because the available evidence did not warrant a claim. The lesson is not that systems should abstain less, but that abstention earns its epistemic function only when users can read it as a judgement about evidence rather than a limit of capability. Second, epistemic access may conflict with proprietary constraints when full provenance, reasoning traces, or model internals cannot be disclosed. In such cases, systems should not be represented as epistemically trustworthy for high-stakes use unless alternative audit mechanisms, contestation pathways, or evidence summaries are provided. Third, resistance to epistemic injustice is difficult to achieve at scale because epistemic recognition requires more than aggregate fairness metrics: it may require corpus audits, deployment-context validation, and engagement with affected communities whose knowledge is often underrepresented in dominant data regimes. Consequently, navigating these tensions while satisfying the three conditions requires deployment-specific design and governance.

\subsection{Non-Fungibility and Its Implications for Evaluation}

Our non-fungibility claim (Section 3.4) draws on the structural logic of plural, threshold-based normative frameworks, including the capabilities approach developed in welfare economics and political philosophy~\cite{nussbaum2011human, sen1999freedom}. In these frameworks, a normative good cannot be reduced to a single averaged or aggregated score; rather, it requires securing multiple constitutive dimensions, each of which protects a distinct and irreducible aspect of that good~\cite{anderson1993value}. Analogously, epistemic trustworthiness should not be evaluated as the weighted sum or numerical average of epistemic humility, epistemic access, and resistance to epistemic injustice. Each condition protects against a structurally distinct pathway through which reliance can lose its epistemic warrant. Strong performance on one dimension therefore cannot compensate for serious failure on another. The conditions should instead be treated as thresholds that must each be satisfied to a level commensurate with the risks, users, and institutional consequences of the deployment context. This non-fungible structure has direct implications for AI evaluation methodology. Composite trustworthiness scores may be useful for summarising broad system performance, but they can obscure failures along individual dimensions. For example, a generative system may score highly on an aggregate rubric because it communicates uncertainty clearly and provides extensive retrieval provenance, while still failing to recognise a specific user group's standpoint as epistemically legitimate~\cite{fricker2007epistemic}. In such cases, the aggregate score may suggest that the system is trustworthy overall, even though reliance remains unwarranted for particular users, communities, or institutional contexts.

To address this limitation, evaluators should use context-sensitive and layer-explicit audit methods that assess each relational condition independently as part of a diagnostic profile. Evaluations of epistemic humility should compare system-side uncertainty signals, retrieval quality, or evidence coverage with the confidence communicated to users, including whether limitation-signalling is maintained across multi-turn interaction. Evaluations of epistemic access should assess whether practitioners can trace the evidence supporting a generative claim, determine whether cited sources actually support the generated text, and understand the output's operational boundaries without disproportionate cognitive or temporal effort. Evaluations of resistance to epistemic injustice should examine whether the pipeline discounts localised domain knowledge, marginalised testimony, or non-dominant language practices, and whether contestation mechanisms allow user expertise to enter the interaction as an active corrective resource.

\subsection{Calibrated Friction as a Design Implication}

A design implication of epistemic trustworthiness is that interfaces should not present all GenAI outputs as epistemically equivalent. When evidence is strong, uncertainty is low, and the stakes of use are limited, systems may support relatively fluent interaction. When evidence is sparse, conflicting, weakly grounded, or consequential for downstream action, the interface should introduce friction that prompts users to inspect, verify, or reconsider the output. We call this \textit{calibrated friction}: interactional friction that increases in proportion to the epistemic risk of relying on a system output. The aim is not to make systems harder to use, but to prevent fluency from becoming unwarranted deference. Prior work shows that friction interventions, such as cognitive forcing functions that require engagement before acceptance, can reduce overreliance without eliminating appropriate reliance~\cite{bucinca2021to}. We reframe this finding normatively: friction is not justified as a behavioural nudge but only when proportionate to genuine epistemic risk. Such friction may take several forms: claim-level uncertainty markers, prompts to inspect cited sources, flags for disagreement across retrieved evidence, or confirmation steps before acting on high-stakes outputs. Across these forms, friction should be grounded in epistemically meaningful reliability signals, such as retrieval quality, semantic uncertainty~\cite{kuhn2023semantic}, or source agreement, rather than in generated self-assessments. The latter are especially risky because verbalised confidence often clusters at high certainty regardless of actual accuracy~\cite{xiong2024can}, thereby reproducing the very miscalibration that friction is meant to counter. Calibrated friction surfaces a central design tension for epistemic agency. Too little friction encourages users to accept outputs that require verification; too much friction becomes paternalistic, obstructing reliance that may be justified. Standard friction interventions primarily address the first risk by slowing users down and prompting additional inspection. Yet this is insufficient in cases of misinformed deference, where users are already attending carefully to the system's expressed reliability signals. In such cases, the problem is not that the user is moving too quickly, but that the interaction directs attention towards the wrong epistemic cue. Friction that prompts further inspection of a poorly calibrated confidence score, a misleading citation, or a fluent explanation may therefore reinforce rather than correct the underlying failure. Calibrated friction must therefore do more than interrupt action. It must redirect users towards cues that support epistemic assessment, rather than merely increasing attention to artefacts that may themselves be misleading. In this sense, calibrated friction operationalises epistemic humility by signalling when reliance should be limited, and epistemic access by creating conditions for users to inspect, question, or withhold reliance before acting. Its relationship to epistemic injustice is more limited but still important. Interface-level friction cannot substitute for data- and model-layer interventions that address representational harms, exclusion, or differential treatment. However, it can play a recognition-supporting role by making contingencies in system behaviour visible, for example by flagging when a response is sensitive to how the user is represented or how the query is framed. Properly designed, friction does not impede epistemic agency; it protects it by turning moments of uncertainty into opportunities for warranted judgement rather than occasions for misplaced deference.

\subsection{Limitations}

Our framework is normative and constitutive rather than implementation-prescriptive: it specifies the conditions an epistemically trustworthy system must satisfy, not the technical means by which they are realised. Mechanisms we discuss illustratively, such as reasoned abstention for humility, contestability for access, and participatory audit for resistance to epistemic injustice, are not fixed proxies or universal prescriptions; their adequacy depends on deployment context, users affected, and institutional stakes. The framework specifies what must be achieved; how it is achieved remains a local, collaborative design problem. Two limitations follow. First, resistance to epistemic injustice is the condition for which deployment-scale operationalisation is least developed: participatory audits and longitudinal studies of recognition patterns are well established in HCI but have not yet been systematically adapted to GenAI evaluation. Second, our cases are drawn from domains where harms are visible and documented; epistemic harms in less legible settings, where miscalibrated deference accumulates without producing discrete failures, are likely underrepresented and require further empirical investigation.

\section{Conclusion}

This paper establishes an account of when human reliance on generative AI is epistemically warranted rather than merely induced by fluency, framing, or interface cues. Grounded in social epistemology, we propose a user--system relational framework with three necessary, non-fungible conditions: epistemic humility, epistemic access, and resistance to epistemic injustice. Real-world cases across high-stakes domains show that hallucination, misgrounded citations, biased rankings, and identity-contingent refusals are not merely output defects but breakdowns in the relational conditions that make reliance warranted. While accuracy, calibration, safety, and fairness remain necessary, they do not capture the interactional conditions under which professional deference forms. We draw two implications. First, evaluation must treat humility, access, and resistance to epistemic injustice as a diagnostic profile showing where and for whom reliance fails, rather than as a composite score. Second, design must support warranted judgement as reliance forms by introducing calibrated friction that is proportionate to epistemic risk and directs users to key reliability cues rather than generated self-assessments. Accordingly, responsible GenAI deployment requires treating these conditions as constitutive requirements and building systems that give users the grounds to inspect, contest, withhold, or appropriately extend reliance on AI outputs.

\bibliography{ref}

\end{document}